\documentclass[10pt,twocolumn,letterpaper]{article}

\usepackage[pagenumbers]{wacv}      

\usepackage{graphicx}
\usepackage{amsmath}
\usepackage{amssymb}
\usepackage{booktabs}
\usepackage{bm}
\usepackage[accsupp]{axessibility}
\usepackage{threeparttable}
\DeclareUnicodeCharacter{0308}{HERE!HERE!}

\usepackage[pagebackref,breaklinks,colorlinks]{hyperref}

\usepackage[capitalize]{cleveref}
\crefname{section}{Sec.}{Secs.}
\Crefname{section}{Section}{Sections}
\Crefname{table}{Table}{Tables}
\crefname{table}{Tab.}{Tabs.}

\def\wacvPaperID{1422} 
\def\confName{WACV}
\def\confYear{2024}

\begin{document}

\title{Prototype Learning for Explainable Brain Age Prediction}

\author{Linde S. Hesse\textsuperscript{1} \quad Nicola K. Dinsdale\textsuperscript{1} \quad Ana I.L. Namburete\textsuperscript{1,2} \\
\textsuperscript{1}OMNI Lab, Department of Computer Science, University of Oxford, UK \\
\textsuperscript{2}Wellcome Centre for Integrative Neuroimaging, FMRIB, University of Oxford, UK \\
{\tt\small \{linde.hesse, nicola.dinsdale, ana.namburete\}@cs.ox.ac.uk}
}
\maketitle

\begin{abstract}
  The lack of explainability of deep learning models limits the adoption of such models in clinical practice. Prototype-based models can provide inherent explainable predictions, but these have predominantly been designed for classification tasks, despite many important tasks in medical imaging being continuous regression problems. Therefore, in this work, we present \textit{ExPeRT}: an explainable prototype-based model specifically designed for regression tasks. Our proposed model makes a sample prediction from the distances to a set of learned prototypes in latent space, using a weighted mean of prototype labels. The distances in latent space are regularized to be relative to label differences, and each of the prototypes can be visualized as a sample from the training set. The image-level distances are further constructed from patch-level distances, in which the patches of both images are structurally matched using optimal transport. This thus provides an example-based explanation with patch-level detail at inference time. We demonstrate our proposed model for brain age prediction on two imaging datasets: adult MR and fetal ultrasound. Our approach achieved state-of-the-art prediction performance while providing insight into the model's reasoning process. 
\end{abstract}

\section{Introduction}
\label{sec:intro}

Deep learning models are typically considered to be black boxes, meaning that it is not possible to understand how a model's prediction was made. This severely limits the adoption of such methods in clinical practice, as the decision-making process needs to be transparent to understand model behavior and gain patients' trust \cite{dinsdale2022challenges}. It is thus vital to develop models that are explainable and, hence, capable of providing insight into their reasoning process \cite{rudin2019stop}. 

The most frequently used methods to explain a model's prediction are saliency-based, which explain the prediction of a trained model \textit{post-hoc} \cite{Velzen2022}. Saliency methods show the importance of each pixel in the input image with regard to the model's prediction. However, these explanations are not always a faithful representation of the original prediction, often resembling edge maps rather than being dependent on the trained model \cite{adebayo2018sanity}. Furthermore, they often result in noisy saliency maps, which are hard to interpret and prone to confirmation bias \cite{alvarez2018robustness,adebayo2018sanity}. On the other hand, inherently explainable models reflect the model's decision-making process by design \cite{rudin2019stop}. However, designing such models for medical imaging is challenging as there is typically a trade-off between performance and explainability. 

An example-based explainable model for the classification of natural images was proposed by \cite{Chen2019}. Their model architecture (\textit{ProtoPNet}) learned a set of embeddings in latent space, referred to as \textit{prototypes}, and used the distances to each of these prototypes to classify a new sample. Each prototype was assigned a class label, and could thus be considered as a representative example (in latent space) for that class. The prototypes in the final model were enforced to equate representations of actual training images, which made it possible to visualize the prototypes. In contrast to \textit{post-hoc} methods, this type of architecture is thus \textbf{inherently explainable}, as the final prediction is directly generated from the prototype distances. However, many important medical image tasks are continuous regression problems, such as brain age prediction \cite{dinsdale2021learning,wyburd2021assessment}. As \textit{ProtoPNet} uses the categorical labels to pull (or push) samples together (or apart), this architecture cannot be directly applied for regression. \looseness -1

In this work, we propose \textit{ExPeRT}: an Explainable Prototype-based model for Regression using optimal Transport. We incorporate metric learning to map the images to an inherently continuous representation space in which the distances between images and prototypes in \textit{latent} space are proportional to their differences in \textit{label} space. This improves upon our earlier work for ordinal regression \cite{Hesse2022}, in which the latent space was not truly continuous, but regularised with an attraction loss that worked on prototypes within a certain set label range.  

The prediction for a new sample is made from a weighted average of prototype labels within a given distance. As all prototypes can be visualized, this provides an intuitive explanation of the prediction for regression tasks. Unlike previous approaches \cite{Chen2019, Hesse2022}, \textit{ExPeRT}'s prototypes are latent representations of whole training images. This is motivated by the fact that the signal of interest in continuous regression problems is often a gradual structural change, rather than class-specific patterns well represented in a single image patch. 

We instead incorporate spatial detail into the model's decision-making process by decomposing each image-level latent distance into patch-wise similarities between image patches. The patch-wise similarities are computed in latent space and then structurally matched using optimal transport (OT). OT finds an optimal matching matrix that contains the \textit{soft assignment} scores between the image patches of the sample and prototype. This matrix is then used to compute a single image-level distance (the Earth Mover's Distance \cite{Rubner2000}). The resulting matches can be inspected to verify whether anatomically corresponding patches are matched together between an image and prototype, providing a detailed spatial decomposition of the image-level distance. \looseness -1

The network is trained using a combination of a \textit{metric loss}, which forces the prototype-image distances to be proportional to label differences, and a \textit{consistency loss} that minimizes the distances between identical image patches under an anatomically justified transformation. Our network and all loss elements are fully differentiable and, thus, the network can be trained end-to-end. 

We demonstrate our approach on the task of brain age prediction for both adult magnetic resonance (MR) and fetal ultrasound (US) images. Brain age prediction is an important medical imaging task: for adult MRI the difference between true and predicted age is a potential biomarker for disease \cite{cole2017predicting, dinsdale2021learning}; during gestation the predicted brain age can be compared to true post-conceptual age to quantify fetal brain development \cite{Namburete2015, everwijn2021association}. We demonstrate that our architecture obtains state-of-the-art performance on both datasets while providing insight into the model's prediction process.

\section{Related Work}





 \textbf{Explainable Brain Age Prediction:} Several studies have attempted to introduce explainability into brain age prediction models, predominantly for adult MRI. Saliency methods have been used to explain brain age predictions \cite{lombardi2021explainable, hofmann2022towards, wyburd2021assessment, li2019deep, Bohle2019}, but their explanations are not always consistent or comparable between different methods \cite{Eitel2019}. Other studies have used patch- or slice-based approaches to predict local brain age \cite{Bintsi2020, Ballester2021}. While these methods provide a more detailed prediction, separate networks have to be trained for each slice or patch, increasing the computational overhead. Using only a single model,  \cite{Popescu2021} used a U-Net to predict brain age voxelwise. The predictions were more fine-grained than slice- or patch-based approaches, but the reported prediction error was considerably higher than the error obtained with baseline models (MAE {\raise.17ex\hbox{$\scriptstyle\sim$}}10 years vs {\raise.17ex\hbox{$\scriptstyle\sim$}}3 years). Alternatively, in \cite{Baumgartner2018, Bass2022}, generative models were used to demonstrate the changes that would be expected in the image for different ages. However, training generative models is challenging and typically requires large amounts of training data which is often unavailable. 



%

 \textbf{Prototype Learning:} \textit{Prototypes} are a set of feature representations in latent space learned during model training, which can be considered representative examples for a certain label. Subsequently, inference for a new sample is performed using the distances to the learned prototypes in latent space, for example by using nearest-neighbour classification \cite{Papernot2018}. In most early deep prototype learning approaches, the learned prototypes were unconstrained points in latent space \cite{Kim2014, Bien2011}, making it challenging to visualise or interpret the learned prototypes. To achieve visualisation of the prototypes, \cite{Li2018} introduced an autoencoder to reconstruct an image from the latent representation. Alternatively, in \cite{Chen2019} it was proposed to constrain the prototypes in the final model to be feature representations of real training images, which could easily be visualized. Further, the prototypes represented patches of training images, as opposed to whole images, creating a more fine-grained explanation. Variations on the \textit{ProtoPNet} architecture have also been applied to several classification \cite{kim2021xprotonet, mohammadjafari2021using, singh2021interpretable, singh2021these,Barnett2021} and ordinal regression \cite{Hesse2022} tasks in medical imaging. 

\begin{figure*}
    \includegraphics[width=1.0\linewidth]{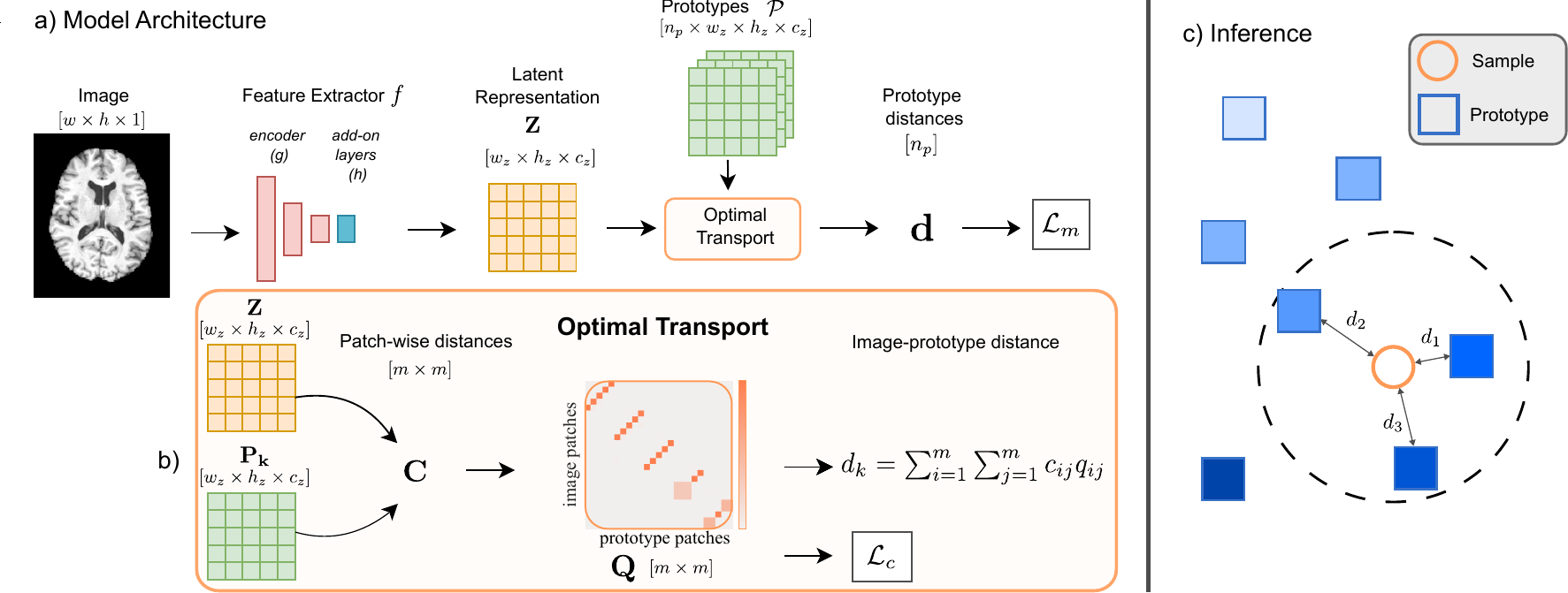}
    \caption{Schematic overview of (a) the \textit{ExPeRT} architecture, (b) distance computation with OT, and (c) inference. The OT matching is shown for one sample and prototype, and is repeated for each prototype to obtain $\bm{d}$. The $m$ is the number of patches per image, given by $w_zh_z$. During inference (c), a sample prediction is made using a weighted average of labels of prototypes within a certain radius.}
\label{fig:overviewfig}
\end{figure*}

\textbf{Deep Metric Learning:} Prototype-based methods are closely related to metric learning approaches. Metric learning aims to learn a feature space where distances between similar samples are small and, inversely, distances between dissimilar samples are large \cite{Lu2017, Zhang2022, Phan2022, Zhao2021, kim2019deep}. The mapping by a neural network is typically achieved by training with contrastive or triplet losses \cite{Schroff2015}. Specifically designed for regression, \cite{Chao2022} regularized the latent space by enforcing the feature distances between samples in a batch to be relative to their difference in label space, measured with the Euclidean distance. However, distances in feature space should be computed along the manifold (\textit{geodesic distances}), and Euclidean distance are only a good approximation for small distances. Thus, samples were weighted by a Gaussian function so as to weigh samples close to each other more than those far apart. Inference was then done using a weighted mean of neighbouring training sample labels. However, this approach requires storing all latent training samples for inference and is dependent on larger batch sizes, which can be problematic when working with large input images due to memory constraints.

\textbf{Optimal Transport (OT):} OT is the mathematical optimization problem computing the shortest distance (or \textit{lowest cost}) between two distributions, given a cost matrix. It can be optimized efficiently when an entropic regularization term \cite{Cuturi2013} is added, and has been incorporated into several deep learning architectures \cite{Torres2021}. Most relatedly, OT has previously been applied to compute distances between pairs of natural images in \cite{Zhang2022, Zhao2021, Phan2022}, where OT was used to match the image patches based on patch-level distances, resulting in improved performance and explainability.





\section{Methods}
The proposed \textit{ExPeRT} model aims to learn a set of prototypes in latent space, each representing a whole image from the training set. Each prototype has a continuous label (e.g. age), and sample predictions are made using a weighted mean of distances to these prototypes. To incorporate patch-level detail, the image-level distances between the image and prototypes are decomposed into patch-level distances and structurally matched using OT matching \cite{Zhang2022, Zhao2021, Phan2022}. A schematic overview of the architecture is shown in Fig. \ref{fig:overviewfig}a. 

\subsection{Network Architecture}
Given an image $\mathbf{X} \in \mathbb{R}^{w \times h \times 1}$, we aim to learn a feature extractor $f$ that can extract a latent representation of $\mathbf{X}$, denoted by $\mathbf{Z}\in \mathbb{R}^{w_z\times h_z \times c_z}$, with $w_z$ and $h_z$ the spatial dimensions and $c_z$ the channel dimension. The feature extractor \textit{f} is composed of a base encoder (\textit{g}), and an additional block (\textit{h}) with two convolutional layers with $1\times1$ kernels and a sigmoid as the last activation. Simultaneously, we learn a set of \textit{prototypes}, which are essentially learned embeddings in latent space, each of the same size as $\mathbf{Z}$: $\mathcal{P} = \{\mathbf{P_i} \in\mathbb{R}^{w_z \times h_z \times c_z}, \forall i \in [1, n_p]\}$, with $n_p$ as the number of prototypes. Each of the prototypes also has an assigned label, resulting in a vector of prototype labels, denoted by $\bm{y^{proto}}$. These prototype labels are assigned uniformly at the beginning of training within the label range present in the training dataset and are fixed during training. The prototypes themselves are considered model parameters, and can, therefore, be trained end-to-end with the feature extractor. For each sample, the network computes the distances to each of the prototypes in latent space, resulting in a vector of distances of length $n_p$, denoted by $\bm{d}$. \looseness -1

\subsection{Prototype Projection} During training, the prototypes are updated with each step and can, therefore, be located throughout the latent space. However, following \cite{Chen2019}, every $N$ epochs these are replaced by the closest latent representation of an image in the training dataset, referred to as the \textit{prototype projection}. Checkpoints are saved only straight after the projection, ensuring that each prototype can be visualized using the corresponding image from the training set.

\subsection{Distance Metric Loss}
The distances between samples and prototypes are regularised, so that for a certain sample with ground-truth label $y$, the distance in latent space to a prototype $k$ is proportional to its difference in label space: $d_k \propto |y^{proto}_{k} - y|$, with $d_k$ and $y^{proto}_k$ as the distance to and label of the $k$th prototype (i.e. the $k$th element of $\bm{d}$ and $\bm{y^{proto}}$), respectively. \looseness -1

However, as the latent space exists on a high-dimensional manifold, computing the feature distances on the manifold between prototype and sample, the geodesic distance, is non-trivial. In this work, we approximate the geodesic distance in the local neighborhood of a sample representation using the Euclidean distance, which is a good approximation for small distances on the manifold \cite{tenenbaum2000global}. 


Building on \cite{Chao2022}, our metric loss regularises distances between samples in the batch and each of the prototypes in the \textit{local neighborhood}. The neighborhood is determined by differences in labels between prototypes and samples, as opposed to the feature distance. The total loss for a single sample  with ground-truth label $y$ is thus given by:

\begin{equation}
    \label{eq:metricloss}
    \mathcal{L}_m(\bm{d}, \bm{y^{proto}}, y) = \sum_{k=1}^{n_p} (|s \cdot d_{k} - (|y^{proto}_k - y|)|) w_k^{train}
\end{equation}

\noindent where $s$ is a learnable parameter scaling the feature distances to the label differences and $w_k^{train}$ a weight of the $k^{th}$ prototype.  $w_k^{train}$ weights the prototype with a Gaussian function based on the label differences, defined by:

\begin{equation}
    \label{eq:metricloss_weight}
    w_k^{train} = e^{-\frac{|y^{proto}_k - y|}{2\sigma^2}}+ \alpha
\end{equation}

\noindent where $\sigma$ controls the size of the neighborhood (i.e. the standard deviation of the Gaussian kernel). $\alpha$ is a small number to prevent the latent embeddings from tangling: without it, samples and prototypes with a large label difference have a negligible effect on each other and could, therefore, be embedded close together in feature space.

\paragraph{Patch-based Distance Metric} In order to train the network with the distance metric loss, distances need to be computed between the sample and each of the prototypes, both of size $h_z \times w_z \times c_z$. A common approach is to use an average pooling operator to compress the spatial dimensions, resulting in a vector of size $c_z$. The Euclidean distance can then be computed between these vectors \cite{Zheng2021, Chao2022}, however, this discards all spatial information. To provide information about the spatial makeup of the distance between a prototype and sample, we propose to instead use a patch-based distance metric. 

The latent representation $\mathbf{Z}$ and a single prototype $\mathbf{P_i}$ can both be considered as sets of $m$ feature vectors: $\{\bm{z_1}, \bm{z_2}, ... \bm{z_m}\}$ and $\{\bm{p_1}, \bm{p_2}, ... \bm{p_m}\}$, with $m = h_zw_z$, and each vector is of size $c_z$. The proposed distance metric computes the Euclidean distances between both sets of feature vectors, $d(\bm{z_i}, \bm{p_j}) \forall i,j \in \{1, ..., m\}$, resulting in a cost matrix $\mathbf{C} \in \mathbb{R}^{m \times m}$. As not all pairwise distances should contribute equally to the image-level distance (i.e. the distance between a patch containing skull and one containing the ventricles is not very meaningful), we use OT to obtain a matching matrix, $\mathbf{Q} \in \mathbb{R}^{m \times m}$ (see below). This matrix $\mathbf{Q}$ can be considered as a \textit{soft assignment} matrix, in which each feature vector $\bm{z_i}$ can be partially matched to more than one feature vector $\bm{p_j}$. These matching and cost matrices can subsequently be used to compute the image-level distance between the sample and the \textit{k}th prototype as follows:

\begin{equation}
    \label{eq:ot_distance}
   d_{k} = \sum_{i=1}^m \sum_{j=1}^m c_{ij}q_{ij}
\end{equation}

\noindent where $c_{ij}$ and $q_{ij}$ are the elements of $\mathbf{C}$ and $\mathbf{Q}$, respectively. 

\paragraph{Optimal Transport (OT)} OT aims to find the matching matrix (or \textit{optimal flow}) that results in the minimal distance (or \textit{lowest cost}) between two distributions. For discrete distributions, given the cost or distance between individual elements, this can be formalized as:
\begin{equation}
    \label{eq:ot_transport}
    \min_{\mathbf{Q}} \sum_{i=1}^m \sum_{j=1}^m c_{ij}q_{ij}
\end{equation}

\noindent which is constrained by the fact that the matching matrix $\mathbf{Q}$ needs to sum up to the initial marginal distributions given by $\bm{w_1}$ and $\bm{w_2}$:

\begin{equation}
    \label{eq:initialdistributions}
    \sum_{i=1}^{m}q_{ij} = \bm{w_1}, 
    \sum_{j=1}^m q_{ij} = \bm{w_2}
\end{equation}

This problem is computationally expensive to solve, and so \cite{Cuturi2013} introduced \textit{entropic regularization} to smooth the optimization problem. The resulting system can efficiently be solved using the classical Sinkhorn divergence algorithm \cite{Sinkhorn1967,Knight2008}. Entropic regularization transforms the optimization of Eq. \ref{eq:ot_transport} into the following optimization problem:

\begin{equation}
    \label{eq:ot_entropy}
    \min_{\mathbf{Q}} \sum_{i=1}^m \sum_{j=1}^m c_{ij}q_{ij} + \epsilon H(\mathbf{Q}) 
\end{equation}

\noindent where $H(\mathbf{Q})$ denotes the entropy function, given by: $H(\mathbf{Q}) = -\sum_{j=i}^m\sum_{j=1}^m q_{ij} log(q_{ij})$, and $\epsilon$ the strength of the entropy regularization. After computing the matching matrix, the minimal distance between the two distributions can simply be computed with Eq. \ref{eq:ot_distance}. 


The initial marginal distributions used in the optimization can be considered as importance scores or weights for each of the feature vectors $\bm{z}_i$ and $\bm{p}_j$. We set both distributions to uniform, but other options could be considered \cite{Zhao2021,Phan2022}. As the OT matching is fully differentiable \cite{Knight2008}, we can train our whole network end-to-end. 

\begin{figure}
    \includegraphics[width=0.48\textwidth]{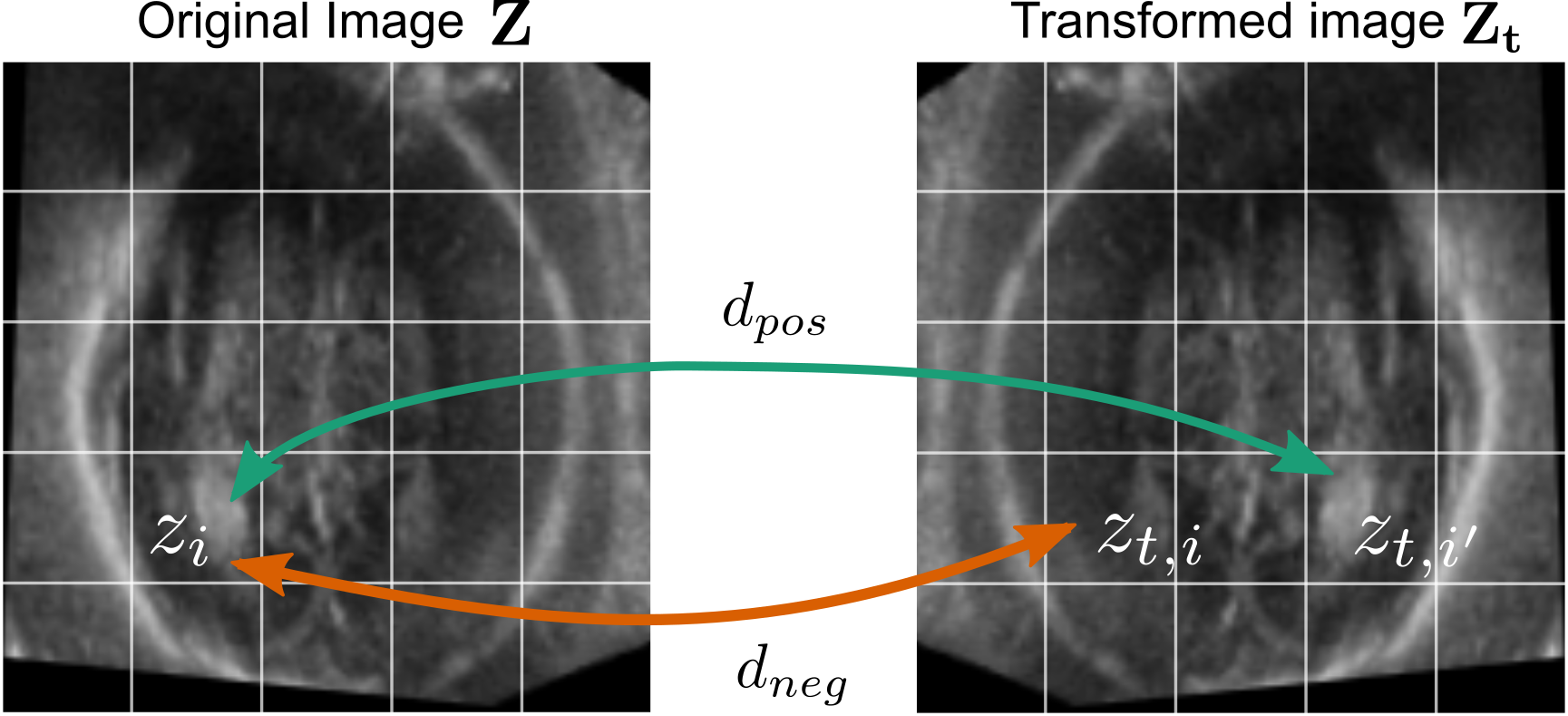}
    \caption{The used consistency loss between a latent image representation ($\mathbf{Z}$) and the representation of a transformed image ($\mathbf{Z_t}$) for a single triplet. The grid indicates the latent pixel size.}
\label{fig:tripletloss}
\end{figure}

\subsection{Consistency Loss}


The OT optimization aims to find structural matches between the image and prototype. To achieve anatomically correct matches, we also introduce a consistency loss. It considers the distances between feature vectors in the latent representation of an image $\bm{z_i} \in \mathbf{Z}=f(\mathbf{X})$ and the feature vectors of the same image under an anatomically justified transformation $T: \bm{z_{t,i}} \in \mathbf{Z_t} = f(T(\mathbf{X}))$ and encourages distances of the same image content, $d(\bm{z_{i}}, \bm{z_{t, i'}})$ with $i'=T(i)$, to be small, while encouraging distances between patches of the same spatial location, $d(\bm{z_{i}}, \bm{z_{t,i}})$, to be large. This resembles a contrastive learning problem where certain feature vectors are pulled together (positive pairs) whereas others are pushed apart (negative pairs). Therefore, our contrastive loss is formulated as a triplet loss, encouraging the distance between an anchor and a positive vector to be larger than the distance between an anchor and a negative vector by a certain margin $\gamma$ \cite{Schroff2015}. To create triplets, we used all vectors $\bm{z_i}$ as anchors, and the respective vectors in $\bm{z_t}$ as positive ($\bm{z_{t,i'}}$) and negative ($\bm{z_{t,i}}$) samples, excluding any triplets for which the positive and negative sample were the same. The Euclidean distance was then computed from each anchor to the positive sample, $d_{pos}$, and the negative sample, $d_{neg}$. The total consistency loss per image computes the sum over all triplets, and can be described by:

\begin{equation}
    \mathcal{L}_c = \sum_{i=1}^{m} max(d(\bm{z_i}, \bm{z_{t,i'}}) - d(\bm{z_i}, \bm{z_{t,i}}) + \gamma, 0)  [ \bm{z_{t,i'}} \neq \bm{z_{t,i}}]
\end{equation}

For training, we used a combination of the distance metric loss (Eq. \ref{eq:metricloss}) and consistency loss as: $\mathcal{L}_{total} = \mathcal{L}_m + \beta \mathcal{L}_c$, where $\beta$ is the weight of the consistency loss.

\subsection{Inference} At inference, the prediction for a new sample is made using a weighted average of the prototype labels within a certain radius $r$ (Fig. \ref{fig:overviewfig}b), given by:

\begin{equation}
    \hat{y} = \frac{\sum_{k=1}^{n_p} w_k^{test} y^{proto}_k}{\sum_{k=1}^{n_p}w_k^{test}}, \quad
    w_k^{test} =  
    \begin{cases}
            e^{-\frac{s \cdot d_{k}}{2(r/3)^2}}, & \text{if  } s \cdot d_{k} \leq r \\
            0, & \text{otherwise}
    \end{cases}
    \label{eq:inference}
\end{equation} 

\noindent in which the weight of each prototype is thus determined by a Gaussian with standard deviation $r/3$.

\section{Experiments}

\subsection{Datasets}
\noindent \textbf{Fetal Ultrasound:} We used 2D ultrasound images sampled from 3D volumes acquired as part of the INTERGROWTH-21\textsuperscript{st} Fetal Growth Longitudinal Study \cite{Papageorghiou2014} to predict gestational age. We used a total of 4290 volumes between 14 and 31 gestational weeks, selected based on having sufficient ultrasound quality \cite{Namburete2015}. All 3D volumes were aligned to the same coordinate system and scaled to the average brain size at 30 GW using an automated alignment method \cite{moser2022bean}. Scaling was performed to enforce the network to learn patterns of structural development rather than only the volumetric size. After alignment, the 2D trans-ventricular plane was extracted from each of the 3D volumes, resulting in a total set of 4290 2D images of size $160\times160$.

\noindent \textbf{Adult MRI:} T1 MRI images from the IXI dataset \cite{IXIdataset} were used for brain age prediction, with ages between 19 and 86. The volumes were preprocessed using the FSL Anat pipeline \cite{FSL}. Only subjects that completed the pipeline successfully were included. For each subject, an axial plane containing the ventricles was selected, as the increase in ventricle size with ageing is well established. This resulted in 561 MR images, each of size $160\times192$. More details for both datasets are given in the supplementary material.

\subsection{Implementation}
We implemented our proposed method with Pytorch Lightning, using PyTorch 1.13 and Python 3.10. All code is publicly available at: \href{https://github.com/lindehesse/ExPeRT_Code}{\url{github.com/lindehesse/ExPeRT_Code}}. Experiments were performed on an A10 GPU with 24 GB RAM and we used an implementation of the Sinkhorn algorithm in logarithmic space to avoid instabilities with training. The Sinkhorn algorithm was run for a maximum of 25 iterations, and the weight of the entropic regularization was set to 0.1 ($\epsilon$ in Eq. \ref{eq:ot_entropy}). We used a ResNet-18 as the base encoder, $g$, pre-trained on ImageNet.

Five-fold cross-validation (train/test 80\%/20\%) was used to tune all hyperparameters, with the reported test performance being the average across folds. 
Networks were trained for 300 epochs for the US data and 150 for the MRI, with a learning rate of $5\times10^{-4}$ and $1\times10^{-4}$, respectively. During training, the images were augmented using small random affine transformations (rotation up to 15 degrees, translation +-6 pixels, scaling between 0.95-1.05). The prototypes were projected to the closest sample in the training set each 25 epochs, starting from 75 epochs. Unless otherwise reported, for both datasets $\alpha$ and $\sigma$ in Eq. \ref{eq:metricloss_weight} were set to 0.05 and 1, respectively, and $r$ in Eq. \ref{eq:inference} to 3; the number of channels in the additional layer block, $h$, and prototypes, $c_z$, to 512, and, the number of prototypes, $n_p$, to 100. The latent space dimensions $w_z$ and $h_z$ were both 5 for the US dataset, and 5 and 6, respectively, for the MRI dataset. During inference, only prototypes within a certain radius $r$ of a sample are considered when generating the final prediction, and it is possible to have no prototypes are within this area, resulting in no prediction. In order to report performance, we assigned the label of the closest prototype to the sample.

Due to shadowing artefacts of the fetal skull, usually only one of the two brain hemispheres is clearly visible in fetal brain US images \cite{Malinger2020}. To match patches in the visible hemispheres with each other, the consistency loss for US was implemented with horizontal flips along the midline of the brain as a geometric transformation (see Fig. \ref{fig:tripletloss}). The weight of the consistency loss $\beta$, was set to 10, and the margin, $\gamma$ to 0.1. For the MR images, the same consistency loss was applied as for the US images.


\subsection{Results}

\begin{figure*}
    \centering
    \includegraphics[width = 0.9 \textwidth]{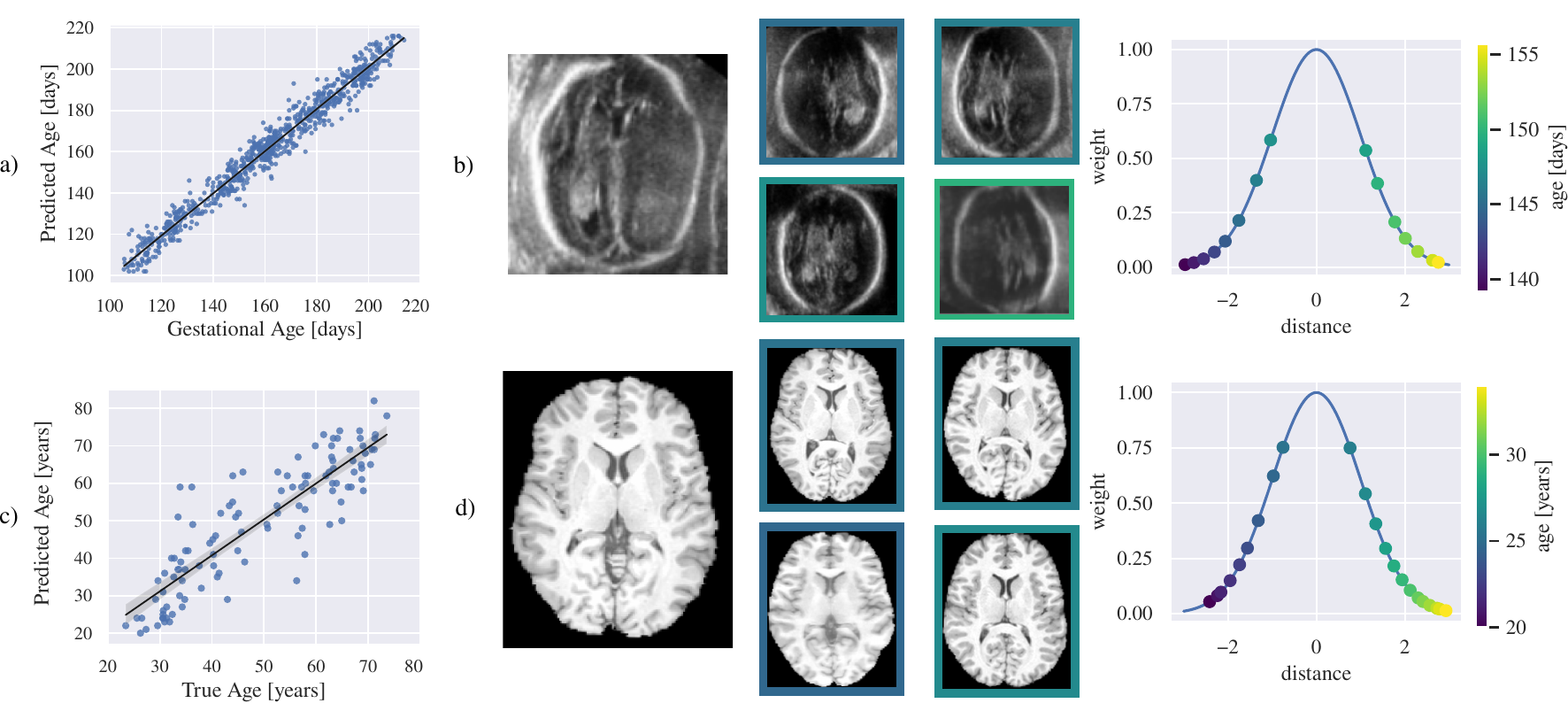}
    \caption{Predicted versus ground-truth age for fetal US (a) and adult MRI (c), as well as an example prediction for US (b) and MRI (d). In b and d, the sample is shown on the left, and the four closest prototypes are shown in the middle. The right-most graphs indicate the weight of each of the prototypes in the final prediction. Prototypes with a label below the predicted value were plotted with negative distances for visualization purposes, but unsigned distances were used in the model itself. For the US image, both the predicted and ground-truth age were 148 days and for the MRI the ground-truth age was 24 years, and the predicted age 25.6 years.}
\label{fig:example_pred}
\end{figure*}

\paragraph{Baseline Comparisons} The quantitative prediction results for both US and MRI are shown in Table \ref{tab:quantitative_results} and in Fig. \ref{fig:example_pred}a and c. We compared our $ExPeRT$ architecture to several non-interpretable age prediction baseline models: the SFCN model + KL loss \cite{peng2021accurate}, winner of the most recent PAC MR brain age prediction challenge; a ResNet-18 + CORAL loss, which showed state-of-the-art age prediction performance for fetal brain age prediction in \cite{lee2023machine}; a vanilla ResNet-18 + MSE loss; and a SFCN + MSE loss, the architecture proposed in \cite{peng2021accurate} but formed as a regression task (as opposed to binning the ages in classes \cite{peng2021accurate}). For both datasets the performance of our method slightly outperforms the baselines, showing that the common assumption that increased explainability results in a decrease in prediction performance does not hold. However, achieving increased prediction performance is not the main aim of this study: rather, we aimed to create a more explainable model without compromising on prediction performance.

Finally, we also compared performance to our previous work \cite{Hesse2022}, an interpretable network proposed for ordinal regression (INSightR-Net). It can be seen that the performance of that model is lower than most other baselines. Furthermore, the learned prototypes did not correspond to age-discriminative image regions. Visualizations of the learned prototypes in INSightR-Net and implementation details for all baselines are given in the supplementary material.

\begin{figure*}
    \centering
    \includegraphics[width = 0.9 \textwidth]{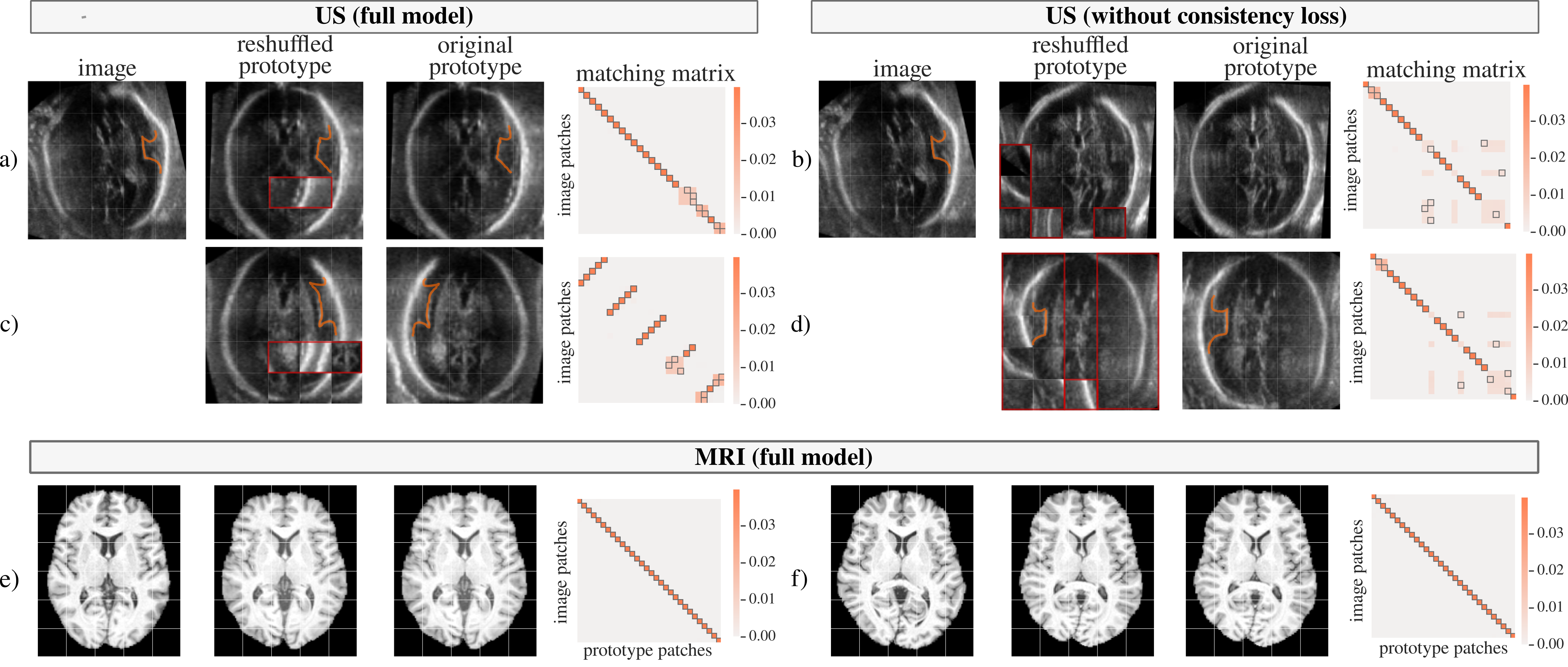}
    \caption{Illustration of OT matching between image sample and prototypes for US (a-d) and MRI (e,f). The obtained OT matching matrix ($\mathbf{Q}$) is given in the last column of each panel, with a black border indicating the most similar prototype patch to each image patch. For each image patch, this most similar prototype patch is visualized in the same location in the \textit{reshuffled prototype}, with patches mirrored if they change side. Patches that are incorrectly matched (according to the known anatomy) have red borders. For US, the matching of the same image sample is shown with two prototypes: one with the same visible hemisphere (a, b), and one with the opposite hemisphere visible (c, d). The full model correctly matches the opposite hemispheres, but the model without consistency loss does not. The Sylvian Fissure has been drawn in orange in each of the visible hemispheres for clarity. For MRI, two examples have been shown for the full model (e,f).}
\label{fig:exampleOT}
\end{figure*}

\paragraph{Example Predictions} Figures \ref{fig:example_pred}b and d show example predictions from our model. The prediction is composed of a weighted average of the labels of prototypes within a certain radius, and these neighboring prototypes are shown next to the sample image. The weight of each prototype is inferred from its distance using a Gaussian function, shown on the right. The color of each point indicates the prototype label, which is also shown with a colored border around each image. In addition to visualizing the prototypes used to make a prediction, our explanation can also decompose the computed distance between each prototype and sample into patch-level matching matrices, which are shown in Fig. \ref{fig:exampleOT}. For each sample-prototype pair, the matching matrix is given in the last column of each panel, indicating which patches are most similar between the prototype and sample. The reshuffled prototype shows the OT matching more intuitively, in which, for each image location, the prototype patch with the highest matching score is shown. In the US dataset, the two visible hemispheres are correctly matched (Fig. \ref{fig:exampleOT}.1a, c) by the full model, illustrating the advantage of using OT matching in our network. 

\begin{table}
    \centering
    \begin{threeparttable}
    \caption{Average MAE ($\downarrow$) across the five cross-validation folds on the test set, with the average standard deviations in brackets. AvgPool represents an ablation of the OT patch-based matching.}
    \label{tab:quantitative_results}
     \begin{tabular*}{\linewidth}{@{\extracolsep{\fill}}lrrrr@{}} \toprule
    
    \multicolumn{1}{c}{} & \multicolumn{2}{c}{Ablat.} & \multicolumn{1}{c}{US} & \multicolumn{1}{c}{MRI} \\ \cmidrule{2-3} \cmidrule{4-5} 
    & $\mathcal{L}_c$ & $h$ & MAE [day]  &  MAE [yr]   \\ 
    \midrule
    Classification \\
    \hspace{1mm} SFCN \cite{peng2021accurate} & - & - & 4.14 (3.31) & 7.26 (5.27) \\
    \hspace{1mm} CORAL  \cite{lee2023machine} & -& -& 4.14 (3.33) & 5.89 (4.96) \\
    \midrule
    Regression \\
    \hspace{1mm} SFCN \cite{peng2021accurate} + MSE  &  - & -  & 5.03 (3.91) &  6.17 (4.30) \\
    \hspace{1mm} ResNet-18 \cite{he2016deep}  & - & - &  4.10 (3.33)   & 5.89 (4.46)  \\
    \midrule
    INSightR-Net \cite{Hesse2022} & - & - & 4.22 (3.41) & 6.43 (5.10) \\
    \midrule
    AvgPool & $\times$ & $\checkmark$ & 4.08 (3.28)    & 5.80 (5.25) \\
    \hspace{1mm} ExPeRT  &$\times$&$\times$ & 4.18 (3.43)  &  6.41 (5.33)  \\
    \hspace{1mm} ExPeRT  & $\times$ & \checkmark& 4.01 (3.27)  & 5.69 (4.59)  \\
    \hspace{1mm} ExPeRT  & \checkmark &$\times$ & 4.16 (3.43)  &  8.02 (6.41)  \\
    ExPeRT & \checkmark & \checkmark& \textbf{3.99} (3.17)  &  \textbf{5.57} (4.51) \\
    
    \bottomrule
    \end{tabular*}
    \label{tab:datasplit}
    \end{threeparttable}
\end{table}


The types of explanations we obtained are very different from more classical explainability approaches, such as saliency methods where a heat map with pixel-level importance scores is generated. We do not aim to compare directly with these kinds of methods but propose our method as an alternative way to increase the explainability of neural networks. Furthermore, while in this study uniform initial distributions were used in Eq. \ref{eq:initialdistributions} (i.e. all patches get the same weight), this could be replaced by other options, such as cross-correlation between the patches \cite{Zhao2021}, to generate importance scores for each of the patches, further improving the explainability of the model.

\begin{figure*}
    \centering
    \includegraphics[width=0.8\linewidth]{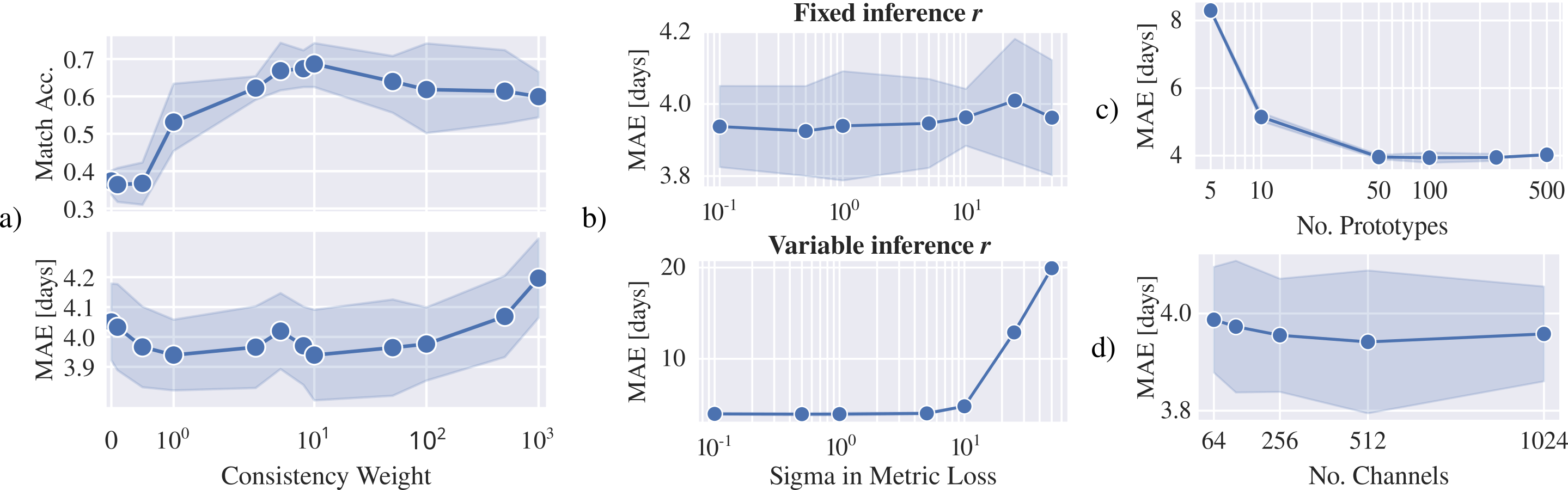}
    \caption{Overview of hyperparameter tuning for the US dataset for the (a) consistency weight, $\beta$, (b) standard deviation in metric loss, $\sigma$, (c) number of prototypes and (d) number of channels in the add-on layers and prototype representations. For $\sigma$ two types of inference are shown, keeping the inference radius, $r$, constant (top) or varying it in line with the training sigma, $r=3\sigma$ (bottom).}
    \label{fig:hyperparametertuning}
\end{figure*}

\paragraph{Ablations} An ablation study on the consistency loss ($\mathcal{L}_c$) and add-on layers ($h$) was completed, and a further ablation study on the OT matching, \textit{AvgPool}, where OT was replaced by pooling the feature representations into a single 1D vector and computing Euclidean distances between these. As the consistency loss works on patch distances, this loss was not applied for the \textit{AvgPool} ablation. All ablation results are shown in Table \ref{tab:quantitative_results}, from which it is evident that for both datasets the add-on layers ($h$) improve the prediction performance. Furthermore, the OT patch-based matching improves performance compared to the \textit{AvgPool} ablation.

The effect of the consistency loss ($\mathcal{L}_c$) is best observed for the US dataset in Fig. \ref{fig:exampleOT},  which shows sample-prototype pairs for the full model trained with (\ref{fig:exampleOT}a,c) and without (\ref{fig:exampleOT}b,d) consistency loss. It can be seen that for the pairs in Fig. \ref{fig:exampleOT}a and b the image and prototype both have a visible right hemisphere (with the Sylvian Fissure annotated), whereas in Fig. \ref{fig:exampleOT}c and d the visible hemisphere of the prototype is opposite to that of the image. The full model correctly identifies the visible hemispheres, matching the patches in these hemispheres with each other. This can also be seen in the reshuffled prototype, in which the sides are flipped. The same pattern of matching between hemispheres was found in all test set volumes. On the other hand, in Fig. \ref{fig:exampleOT}d it is evident that without consistency loss, no matching occurs between the two opposite visible hemispheres. Instead, the same spatial location in both the image and prototype are matched, as shown by the diagonal matching matrix. This illustrates that the consistency loss is responsible for matching the correct hemispheres. In this work only flipping across the midline of the brain has been used as geometric transformation in the consistency loss: future work should consider other transformations, based on the geometric variation present in the dataset. It is also important to note that no hemisphere labels were used during training as the consistency loss is unsupervised.  \looseness -1

In brain MR images, both hemispheres are visible, and there is thus less need to enforce inter-hemispheric patch similarity with a consistency loss. For this reason, the effect of this loss component for the MR images is less pronounced, with only subtle differences between the models trained with or without consistency loss. For both models, the matching matrices found (Fig. \ref{fig:exampleOT}e,f) are mostly according to the diagonal, suggesting that each patch in the image is matched with the same spatial location in the prototype. As our images were aligned to the same template space (MNI), this matching is expected and shows that the model can learn the similarity between corresponding patches. 



%

\paragraph{Sensitivity to Hyperparameter Choice} The average validation performance for each of the important hyperparameters is shown in Fig. \ref{fig:hyperparametertuning} for the US dataset. For the consistency weight (Fig. \ref{fig:hyperparametertuning}a), the top plot shows the patch-matching accuracy, which is computed from the overlap with approximated ground-truth matching matrices determined from the known visible hemisphere in each sample. It is evident that increasing the consistency weight results in an improvement of both the MAE and patch-matching accuracy up to a weight of 10, after which it starts deteriorating again. It should be noted that the consistency loss is about four orders of magnitudes smaller than the metric loss during training, hence the small effect for low weights.

In Fig. \ref{fig:hyperparametertuning}b the standard deviation ($\sigma$) in the metric loss (Eq. \ref{eq:metricloss_weight}) is varied. The top and bottom panels show the results of the same training runs, but during inference, the radius is either fixed ($r=3$, top panel) or adapted based on the $\sigma$ during training ($r = 3\sigma$, bottom panel). The training $\sigma$ has only a small effect on performance throughout the range of values tried, whereas the inference radius does considerably affect the performance, most notably at higher values. This is beneficial as the inference $r$ can be easily adjusted after training, and can thus be optimized offline.

The number of prototypes and the number of channels in the prototype representations (and in the add-on layers) are shown in Fig. \ref{fig:hyperparametertuning}c and d respectively. Both show improved performance when increasing the number of prototypes or channels, leveling off for higher values. 

Overall, these results show that the hyperparameters introduced in our model behave as expected, confirming the stability of our approach and showing that the results are not overly sensitive to the hyperparameter selection. 
\label{sect:hyperparametertuning}

\section{Conclusion}
We have presented ExPeRT, a novel explainable model for continuous regression, based on prototype learning and patch-based OT matching. Our model achieves competitive prediction performance on two brain age prediction datasets: fetal US and adult MRI. For both datasets, the patch-based distance metric was able to correctly learn the structural matches between the sample image and prototypes. Our approach is versatile and can be applied to other continuous regression problems, beyond medical imaging.

\section*{Acknowledgements}
L.S.H. acknowledges the support from the \mbox{EPSRC} Doctoral Prize award. A.I.L.N. and N.K.D. are supported by the Bill and Melinda Gates Foundation.
{\small
\bibliographystyle{ieee_fullname}
\bibliography{egbib}
}

\end{document}


\date{Paper ID 1422}
\maketitle
\section{Dataset Details}

\paragraph{Fetal Ultrasound} The US volumes used in this study were from the \\ INTERGROWTH-21\textsuperscript{st} Fetal Growth Longitudinal Study \cite{Papageorghiou2014}. That study aimed to describe fetal growth and neurodevelopment in a geographically diverse, healthy population of women, enrolled before 14 weeks' gestation. The volumes were acquired with a Philips US device (Philips HD-9, Philips Ultrasound, USA) using a curvilinear abdominal transducer. All gestational ages were determined based on the last menstrual period and confirmed with US crown-rump measurements that needed to agree within 7 days.                                     
\paragraph{Adult MRI} The adult MR images used in this study were from the IXI dataset, which contains MR images from healthy subjects. The images were acquired across three imaging sites in London, UK, on different imaging systems. We preprocessed all image volumes using the FSL Anat pipeline, the key stages of which are skull stripping, bias field correction, and linear registration to the 1mm MNI template. 

\section{Baseline Implementations}
For all baseline implementations, the learning rate was tuned individually from the following options: 1e-3, 5e-4, 1e-4, 5e-5, 1e-5. 

\paragraph{SFCN} The SFCN architecture \cite{peng2021accurate} is a classification-based method that is trained for regression with a KL-divergence loss between a Gaussian distribution around the true age, and the predicted class probabilities. The number of days/years per bin was selected using cross-validation based on the validation performance, and selected from (1, 2, and 7 days per week) for US, and from (1, and 2 years per bin) for MRI. For both datasets, the smallest number of days/years per bin resulted in optimal performance and was therefore reported in the main results section. This corresponded to the original SFCN implementation where 1 year per bin was used. The standard deviation of the Gaussian distribution was set to 1 day/year, following \cite{peng2021accurate}.

\paragraph{SFCN-reg} This model uses the same backbone architecture as proposed in \cite{peng2021accurate} but with the last classification layer replaced by a layer with a single regression output. This model was then trained with a mean squared error loss. 

\paragraph{CORAL} The model referred to as CORAL in the main text consists of a classification ResNet-18 with a CORAL ordinal regression loss, which has been applied for fetal brain age prediction \cite{lee2023machine}. In the original work, a bin size of 1 week was used, but we selected the optimal bin size from the same options as for the SFCN architecture. Also for this architecture, the optimal bin size was 1 day/year per bin, and this value has therefore been reported in the main paper.

\paragraph{INSightR-Net} We introduced this model in earlier work \cite{Hesse2022} and implemented it according to the publicly available code at \footnote{\url{https://github.com/lindehesse/INSightR-Net}}. For an equal comparison, the same number of prototypes was used as in the proposed \textit{ExPeRT}, 100, with 512 channels in the add-on layers. The weights of the loss components were kept equal to the original implementation ($\alpha_{MSE}=1$, $\alpha_{Clst}=1$, $\alpha_{PSD}=10$), and the $\Delta_l$ in the $\mathcal{L}_{Clst}$ to 2 days/years (due to the larger label range for the age prediction, the original paper used 0.5).

\newpage

\section{Additional Results INSightR-Net Baseline}

\begin{figure*}[h!]
    \includegraphics[width=1.0\linewidth]{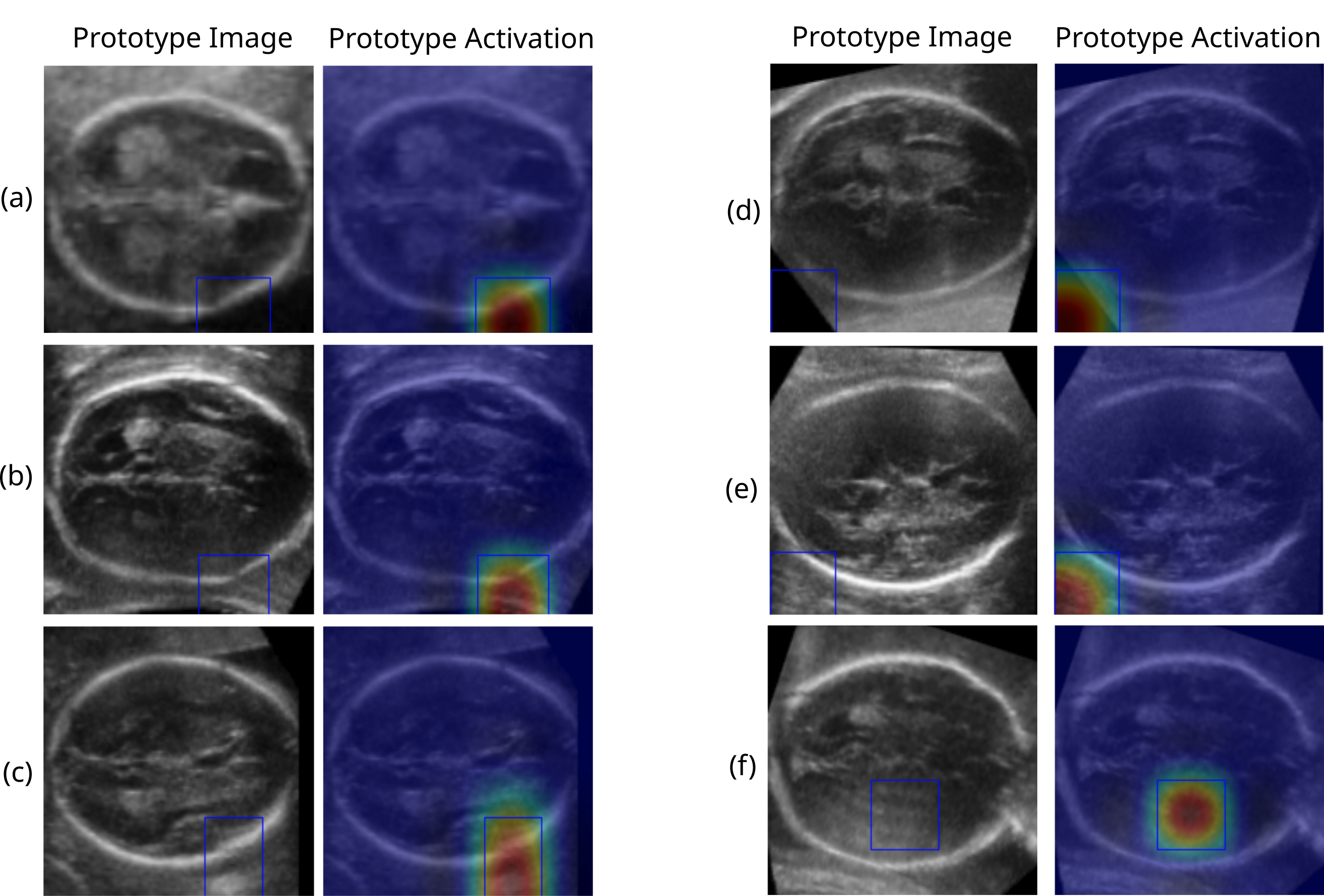}
    \caption{Prototypes recovered from baseline INSightR-Net \cite{Hesse2022} The shown prototypes were randomly picked from a single trained model.  It can be seen that the prototype patches (indicated by the blue boxes) do not recover specific age-discriminative regions but rather background and skull regions.}
\label{fig:overviewfig}
\end{figure*}

\newpage
{\small
\bibliographystyle{ieee_fullname}
\bibliography{egbib}
}